\documentclass[conference]{IEEEtran}
\IEEEoverridecommandlockouts

\usepackage{cite}
\usepackage{amsmath,amssymb,amsfonts}

\usepackage{algorithm}
\usepackage{algpseudocode}

\usepackage{graphicx}
\usepackage{textcomp}
\usepackage{xcolor}
\usepackage{multirow}
\usepackage{soul}
\usepackage{hyperref}
\def\BibTeX{{\rm B\kern-.05em{\sc i\kern-.025em b}\kern-.08em
    T\kern-.1667em\lower.7ex\hbox{E}\kern-.125emX}}
\begin{document}

\title{Towards Resilience: Slice-Driven Orchestration of Mobile IoMT in 6G-Edge-Continuum\\
\thanks{This work was supported by the 6G Flagship (Grant Number 369116) and 6GESS (Profi6 Grant Number 336449) funded by the Research Council of Finland.}
}

\author{\IEEEauthorblockN{Muhammad Zeeshan}
\IEEEauthorblockA{\textit{Centre for Wireless Communications} \\
\textit{University of Oulu}\\
Oulu, Finland \\
muhammad.zeeshan@student.oulu.fi}
\and
\IEEEauthorblockN{Hafiz Faheem Shahid}
\IEEEauthorblockA{\textit{Centre for Wireless Communication} \\
\textit{University of Oulu}\\
Oulu, Finland \\
faheem.shahid@oulu.fi}
\and
\IEEEauthorblockN{Erkki Harjula}
\IEEEauthorblockA{\textit{Centre for Wireless Communication} \\
\textit{University of Oulu}\\
Oulu, Finland \\
erkki.harjula@oulu.fi}
}

\maketitle

\begin{abstract}
Emerging healthcare services, such as real-time diagnostics, remote health monitoring and computer-assisted surgery, require stringent Quality of Service (QoS) guarantees while operating under limited and heterogeneous edge resources. Existing Internet of Medical Things (IoMT) orchestration methods primarily focus on communication metrics such as latency and bandwidth, while overlooking edge resource constraints and dynamic adaptation to workload and user mobility. This paper presents a novel resource-slicing-based adaptive orchestration framework that integrates 5G network slicing (NS), resource-aware orchestration, and mobility-aware task offloading into a single adaptive system. IoMT services are assigned to distinct 5G network slices while jointly considering CPU capacity, memory availability and amount of storage, enabling more efficient service placement and execution. In addition, the system supports user mobility and dynamic task offloading to maintain service continuity under changing network conditions. SimPy-based simulation demonstrates that the proposed method enhances service deployment efficiency, reduces request blocking, and improves user acceptance while maintaining reliable performance across diverse IoMT scenarios.

\end{abstract}

\begin{IEEEkeywords}
Network Slicing, Service Orchestration, Quality of Service, Edge-Cloud Continuum, 6G.
\end{IEEEkeywords}

\section{Introduction}
The increasing adoption of Internet of Medical Things (IoMT) technologies is transforming modern healthcare services through applications, such as computer-assisted surgery, wearable monitoring, telemedicine and diagnostic imaging. Different IoMT applications produce heterogeneous workloads with diverse Quality of Service (QoS) requirements, ranging from ultra-low latency for computer-assisted surgery to massive connectivity for patient monitoring and high bandwidth for medical imaging. Efficient service management remains a significant challenge in distributed computing systems \cite{islam2025hybrid}.

Traditional cloud-centric IoMT applications suffer from high latency and are highly dependent on reliable internet connection. Delays can negatively affect time-critical medical services, while failing Internet connection may completely block the service \cite{aguru2022integrated}. While edge computing brings computation and storage resources closer to users end to improve responsiveness, edge nodes usually operate under limited and heterogeneous resource availability. In practical healthcare environments, changing network conditions, dynamic workloads, and user mobility further complicate service orchestration and resource management across distributed edge environments.

Network Slicing (NS) supports heterogeneous IoMT workloads with guaranteed QoS over a single physical infrastructure in 5G and beyond networks. In healthcare scenarios, ultra-Reliable Low-Latency Communication (uRLLC) slice supports critical-care services, massive Machine-Type Communication (mMTC) slice enables large-scale wearable patient monitoring, and enhanced Mobile Broadband (eMBB) slice supports high-throughput applications including real-time diagnostic imaging \cite{11524197}. Combining NS with edge computing enables allocation of communication and computational resources for IoMT applications more efficiently\cite{shahid2024resource}.

Existing orchestration approaches mainly consider communication-level QoS metrics such as latency and bandwidth, while giving limited attention to computational resource constraints, user mobility, and adaptive service deployment across distributed edge nodes. These gaps affect service continuity and reliability for life-critical IoMT applications, highlighting the need for an adaptive orchestration framework capable of jointly managing network slicing, computational resource allocation, load balancing and mobility-aware task offloading in distributed edge environments. However, a comprehensive IoT service orchestration architecture in a multi-tier computing continuum, including IoT-cloud, MEC, fog computing, and edge-cloud computing, has been analysed in \cite{11392773}.

The main contribution of this paper is the development of a resource-slicing-based adaptive orchestration framework that integrates 5G network slicing (NS) with resource-aware orchestration \cite{10503485} to support diverse IoMT QoS requirements. In addition, the proposed framework incorporates mobility-aware task offloading to maintain service continuity under user movement and changing network conditions. Finally, SimPy-based simulation demonstrates improvements in service deployment efficiency, resource utilization, and application acceptance ratio compared to a baseline approach. The remainder of this paper is organized as follows: Section II reviews related work, Section III presents the proposed framework, Section IV discusses results, and Section V concludes the paper.

\section{RELATED WORK}

\subsection{Network Slicing in 5G Networks}
NS is a key enabling technology in 5G networks, allowing the same physical infrastructure to be divided into logical slices, configured according to the latency, reliability, and bandwidth needs \cite{shahid2024resource}. In healthcare systems, this capability is important because IoMT applications generate highly diverse traffic patterns. For instance, emergency healthcare services require very low latency and reliable communication, whereas wearable monitoring devices mainly require support for massive connectivity. Similarly, medical imaging and remote consultation services depend on high data throughput. As the number of connected IoMT devices increases, maintaining stable QoS and efficient utilization of network resources requires intelligent and adaptive slice management mechanisms \cite{suleiman2023telemedicine}.

Previous studies have explored different resource allocation approaches for network slicing in 5G systems. Reservation-based methods provide stronger isolation between slices and improve service reliability, but they may reduce overall resource efficiency because resources remain reserved even during low utilization periods. In contrast, share-based approaches improve utilization by allowing statistical sharing of resources among slices, although this can weaken isolation and affect QoS guarantees during congestion \cite{9272281}. Loven et al. in \cite{10287519} introduce the concept of AI-driven semantic slicing to optimize the computational resource utilization to deploy IoT applications in the computing continuum with 6G networks.  

However, most existing studies mainly focus on communication-level QoS parameters such as bandwidth and latency, with limited attention to joint management of network and computational resources. As a result, there is still a need for orchestration mechanisms that can efficiently coordinate network slicing with edge resource management for IoMT applications.

\subsection{ Resource Slicing in Edge Computing}
Edge nodes operate with limited processing, memory, and storage capacity, making efficient resource management a critical task for latency-sensitive IoMT applications. Resource slicing (RS) addresses this issue by partitioning network and compute resources into dedicated virtual slices, allocated to specific applications based on their requirements \cite{husain2018mobile}. Recent research has explored combining resource slicing with network slicing in MEC-enabled 5G environments to improve overall system efficiency. These approaches aim to coordinate communication and computational resource allocation so that heterogeneous IoMT workloads can be supported more effectively. While such methods improve scalability and resource utilization, they still face challenges related to dynamic workloads, mobility support, and constrained edge resources  \cite{bahramisirat2024multi}.

Some studies further propose microservice-based architectures and task offloading strategies to enhance service responsiveness by processing data closer to the edge. This reduces end-to-end latency and enables more flexible utilization of CPU, memory, and bandwidth resources across distributed nodes \cite{hwang2021iot}.
Despite this, most approaches are designed for generic IoT systems and do not fully address the requirements of adaptive orchestration for heterogeneous IoMT services. Joint management of network slicing, resource constraints, and mobility-aware service execution in edge environments remains an open problem.

\subsection{Intelligent Edge Orchestration and Load Balancing}

Static resource allocation policies are insufficient for dynamic IoMT workloads, motivating the use of intelligent dynamic orchestration and load balancing in edge computing environments. Existing approaches use AI-based orchestration to distribute workloads between edge and cloud, where delay-sensitive tasks are processed at the edge and compute-intensive tasks are deployed to the cloud data centers, ensuring latency, energy efficiency, and service performance \cite{11220483}. Recent works also integrate Machine Learning (ML) with Software-Defined Networking (SDN) and NS to enable adaptive node selection and dynamic resource management under changing network conditions. Reinforcement learning methods further improve adaptability by learning optimal allocation policies from historical system states, supporting efficient operation in multi-tenant environments \cite{natalino2018machine}.

In parallel, load balancing techniques are used to avoid resource overloading and improve utilization in distributed edge systems. AI-based methods such as swarm intelligence and PSO enable dynamic task distribution across nodes, reducing delay and improving system efficiency \cite{shahidswarm}. Further studies show that distributed and dynamic load balancing strategies consistently outperform static methods for edge environments \cite{pydi2020analytical}.
However, most of the existing solutions depend on complex model training, limiting their suitability for real-time IoMT deployments under constrained edge resources. This highlights the need for lightweight and adaptive orchestration  approach capable of jointly managing network slicing, resource constraints, and mobility without training dependency.

In summary, existing works address network slicing, resource management, and intelligent orchestration in isolation, without jointly considering all three alongside user mobility in IoMT edge environments.

\section{Proposed System Model}

\subsection{System Architecture}
\begin{figure}[!ht]
\centerline{\includegraphics[width=1\linewidth, height =8cm]{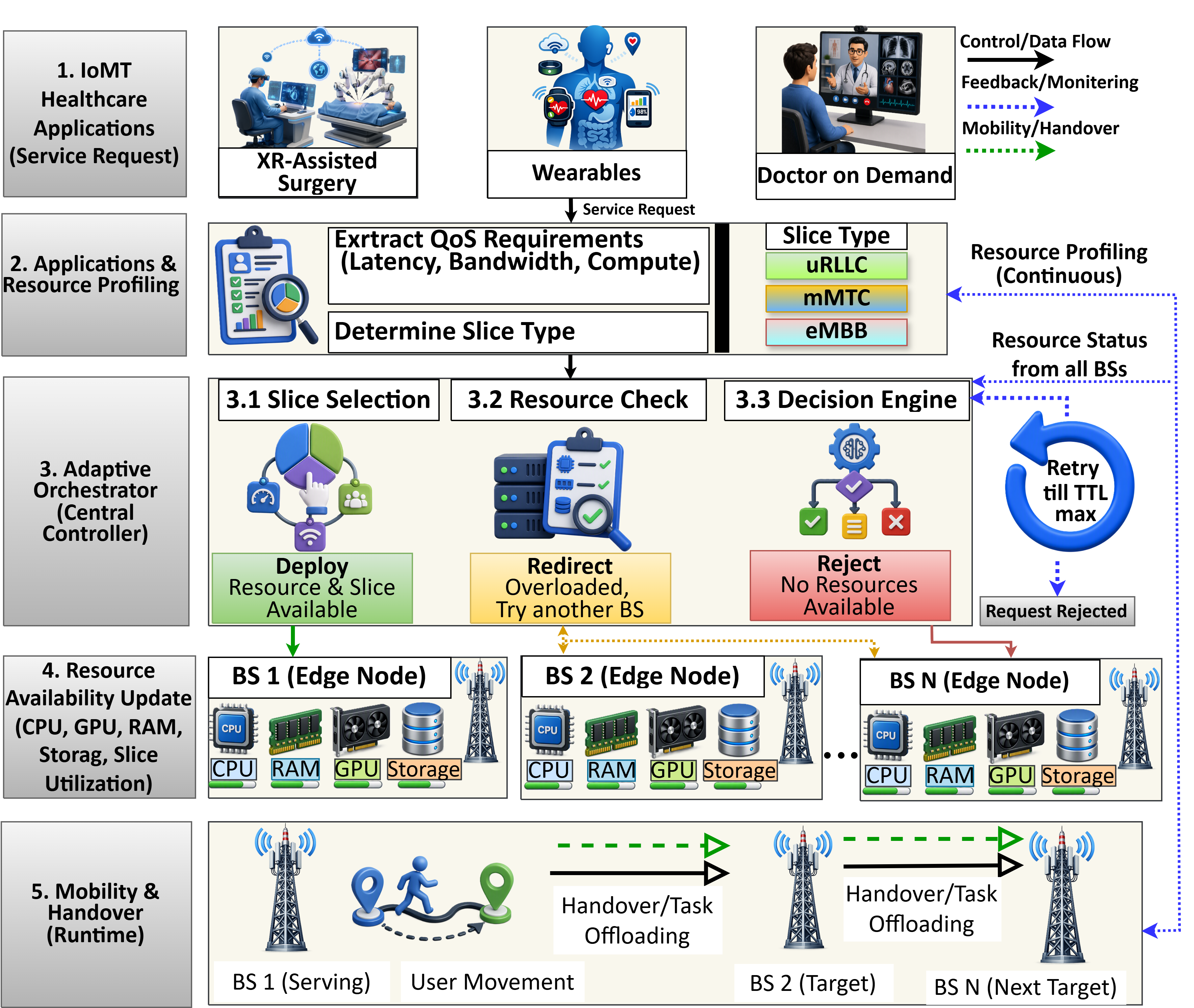}}
\caption{ System architecture for IoMT application orchestration.}
\label{Flow chart of propsoed system}
\end{figure}

The proposed system model combines 5G network slicing with distributed edge infrastructure to support adaptive orchestration of heterogeneous IoMT services. It is designed to handle heterogeneous healthcare services with different computation, latency, and reliability requirements under dynamic network conditions and user mobility. The adaptive orchestration framework jointly considers application requirements and real-time resource availability at each BS. Each IoMT application is characterized through application profiling, which defines its QoS, latency, bandwidth, and computational needs. Each BS continuously reports its available resources through resource profiling, capturing the current state of compute and network capacity. Based on these profiles, service requests are mapped to an appropriate BS and NS that is uRLLC, mMTC, or eMBB. If the nearest BS is overloaded, the orchestrator dynamically redirects the request toward an alternative BS in the same or adjacent coverage region to prevent service degradation and balance load across the network as shown in Fig.~\ref{Flow chart of propsoed system}.  

To support mobility active sessions are continuously monitored. When a user transitions to a new BS, a handover process is initiated and ongoing services are offloaded to another BS to maintain service continuity. When a specific slice becomes saturated, unused resources from other slices can be temporarily reallocated to maintain service continuity for high-priority applications, reducing request blocking and improving overall system efficiency.

\subsection{IoMT Use Cases}
Different IoMT applications are categorized based on their bandwidth, latency, and computational requirements, which enable orchestration based on NS and resource awareness within edge computing environments.

\paragraph{XR-assisted Surgery} XR-assisted surgery (XS) enables real-time 3D visualization and interaction during surgical procedures. To support reliable operation with strict latency requirements below 20 ms, it is assigned to the uRLLC slice. Substantial processing capacity is required for real-time 3D model rendering  and sensor-data processing. Depending on the quality of imaging and XR streams, bandwidth requirements varies between 10 Mbps and 100 Mbps \cite{selem2019health}.

\paragraph{Wearable-based Health Monitoring} Wearable health monitoring (WHM) involves in continuous collection of patient data such as heart rate, SpO2, and glucose levels from wearable sensors. It generates low-rate but frequent transmissions from a large number of devices and is mapped to the mMTC slice. The application requires scalable connectivity with lightweight edge processing. Typically, bandwidth usage remains in range of 1-2 Mbps per device, while delays up to 100 ms are acceptable for continuous monitoring \cite{ajakwe2022key}.

\paragraph{Doctor-on-Demand} Doctor on Demand (DoD) provides real time audio and video-based medical consultations between patients and healthcare professionals.  it is assigned to the eMBB slice due to its requirement for high BW and stable communication. Bandwidth demand typically varies between 10 and 150 Mbps, with an acceptable latency threshold of 150 ms to maintain smooth and interactive communication \cite{selem2019health}.

\subsection{Orchestration Algorithm and Workflow}

Algorithm \ref{alg:iomt} describes the deployment procedure for IoMT applications across multiple BSs among different regions. For each application the algorithm identifies a suitable NS and a suitable BS, verifies computational and slice-level resource availability and decides to deploy, offload, or reject the application based on current network conditions.

\algnewcommand\algorithmicinput{\textbf{Input:}}
\algnewcommand\algorithmicoutput{\textbf{Output:}}
\algnewcommand\Input{\item[\algorithmicinput]}
\algnewcommand\Output{\item[\algorithmicoutput]}

\begin{algorithm}[ht]
\caption{Resource-aware IoMT Service Orchestration in Distributed Base Station Networks}
\footnotesize
\label{alg:iomt}
\begin{algorithmic}[1]
\Input
\Statex \hspace{1em} Service set $\mathcal{W} = \{\text{XS}, \text{WHM}, \text{DoD}\}$
\Statex \hspace{1em} Slice set $S = \{\text{uRLLC}, \text{mMTC}, \text{eMBB}\}$
\Statex \hspace{1em} Base Stations $BS = \{bs_0,\dots,bs_X\}$ with
\Statex \hspace{1em} $\text{Capacity}(bs) = \{CPU, RAM, Stor, GPU\}$
\Statex \hspace{1em} Coverage area $= 2000 \times 2000\,\text{m}^2$,\quad 
\Statex \hspace{1em} $T_{run} = 100\,\text{s}$,\quad $TTL = Max_{attempts}$
\Statex
\State Get resources from BS $Res(bs) \gets \text{Capacity}(bs)$, $\forall\, bs \in BS$
\Statex
\For{$t \gets 0$ \textbf{to} $T_{run}$}
    \For{each service $w \in \mathcal{W}$}
        \State $N_{attempts} \gets 0$
        \State $s \gets \text{AssignSlice}(w)$
        \While{$N_{attempts} < Max_{attempts}$}
            \State $bs^{*} \gets \text{NearestBaseStation}(w, t)$
            \If{$Res(bs^{*}) \geq \text{ResourceDemands}(w)$ \textbf{and}
                $\text{CheckSliceAvailability}(bs^{*}, s)$}
                \State Assign $w \rightarrow (bs^{*}, s)$
                \State Iterate $Res(bs^{*})$, $\text{SliceCap}(bs^{*}, s)$
                \State \textbf{break}
            \ElsIf{ Alternate $bs \in BS$ available}
                \State Redirect $w \rightarrow (bs, s)$
                \State \textbf{break}
            \Else
                \State $N_{attempts} \gets N_{attempts} + 1$
            \EndIf
        \EndWhile
        \If{$N_{attempts} = Max_{attempts}$}
            \State Reject $w$
        \EndIf
    \EndFor
\EndFor
\Statex
\Output
\Statex \hspace{1em} Deployment set $D$ and metrics $\{deployed,\, rejected,\, redirected\}$
\end{algorithmic}
\end{algorithm}

\subsection{Application and Resource Profiling}
Efficient orchestration of IoMT applications relies on two profiling mechanisms. Application profiling captures the requirements of each IoMT application in terms of latency, bandwidth, and computational needs (CPU, GPU, RAM, storage) and maps them to the appropriate 5G network slice as shown in Table \ref{IoMT application resource requirements}.
Resource profiling tracks real-time compute and communication availability at every BS, maintaining information about its remaining resources and slice-specific bandwidth.

Using both profiles, the orchestrator performs real-time admission and placement decisions. An IoMT application is deployed at a selected BS only if sufficient computational resources are available and the associated network slice can satisfy its QoS constraints. If the nearest BS does not meet these conditions, the orchestrator redirects the request to an alternative BS within the same or neighbouring region. In case of persistent resource scarcity, the request may be delayed or rejected to preserve service quality for running applications.

\subsection{Resource Slicing for IoMT Services}
NS in the proposed IoMT framework enables logical separation of the physical network into multiple virtual network slices, each designed to meet specific service requirements. This isolation ensures that heterogeneous applications can run simultaneously without affecting system performance, while maintaining dedicated QoS guarantees.
Three main NSs are considered: uRLLC for life-critical services such as XR-assisted surgery, mMTC for large-scale wearable-based monitoring, and eMBB for bandwidth-intensive applications like remote consultation. Each NS is configured with specific latency, bandwidth, and reliability constraints based on the application. The orchestrator uses slice-aware policies jointly with compute resource availability (CPU, RAM, GPU, storage) to assign each application to the most suitable BS and NS.

\begin{table}[htbp]
\centering
\caption{Resource requirements and network slice mapping for IoMT services.}
\renewcommand{\arraystretch}{1.4}
\setlength{\tabcolsep}{2.5pt}
\small
\resizebox{\columnwidth}{!}{%
\begin{tabular}{|c|c|c|c|c|c|c|c|}
\hline
\textbf{Service} & \textbf{CPU} & \textbf{GPU} & \textbf{RAM} & \textbf{Storage} & \textbf{BW} & \textbf{Latency} & \textbf{Slice} \\
& \textit{(Cores)} & \textit{(Cores)} & \textit{(MB)} & \textit{(MB)} & \textit{(Mbps)} & \textit{(ms)} & \\
\hline
Doctor on Demand           & 6 & 1--4 & 4096 & 2048 & 10--150 & $\leq$150 & eMBB  \\
\hline
XR-assisted Surgery        & 8 & 1--8 & 8192 & 4096 & 10--100 & $\leq$20  & uRLLC \\
\hline
Wearable Monitoring & 2 & --   & 1024 & 512  & 1--2    & $\leq$100 & mMTC  \\
\hline
\end{tabular}%
}
\label{IoMT application resource requirements}
\end{table}

\subsection{Proposed System Configuration and Experimental Setup}
\begin{figure}[!ht]
\centerline{\includegraphics[width=1\linewidth, height =8cm]{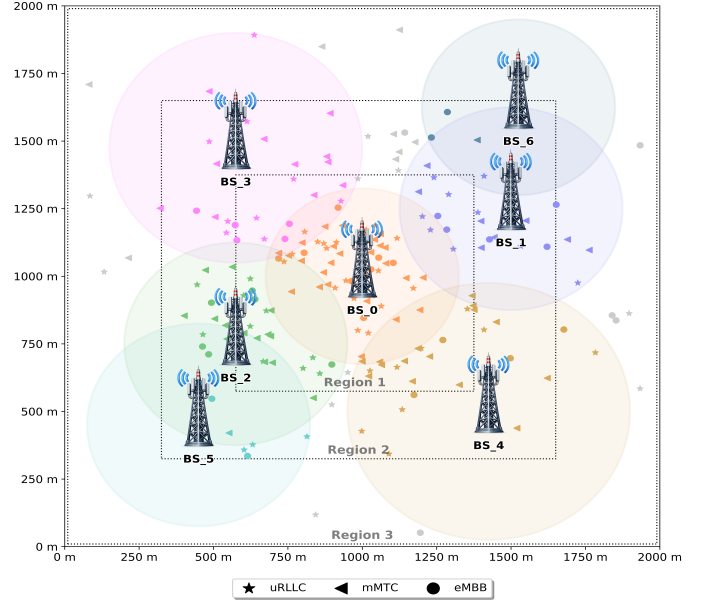}}
\caption{ IoMT users association with nearest base stations across regions in the proposed system.}
\label{simulation map}
\end{figure}

The simulation environment is designed in SimPy to evaluate the proposed IoMT orchestration framework in a distributed 5G edge network. The scenario consists of 7 base stations in 2000 by 2000 meters area as shown in Fig.~\ref{simulation map}, where 500 users move randomly across 3 regions runs for 100 s. The total combined resource capacity across the 7 BSs is: 2300 CPU cores, 2200 GB RAM, 1400 GPU cores, and 1400 GB storage. Each base station has a circular coverage zone. Users inside zone remain connected, while users outside it lose service until they re-enter the BS coverage area.
User mobility triggers frequent handovers between base stations, which is used to assess service continuity, especially for delay-sensitive IoMT applications. Each BS is provisioned with distinct computational capacities, reflecting realistic edge deployments, where resources are inherently heterogeneous. Performance is evaluated using metrics such as service acceptance rate, blocking rate, handover frequency, and resource utilization.

\subsubsection{Proposed System}
The total available resources across the network are distributed among the base stations, including CPU cores, RAM, GPU capacity, and storage, and are used by the orchestrator for runtime scheduling and offloading decisions.
Applications are mapped to appropriate slices based on their QoS requirements. As users move across regions, the orchestrator dynamically handles handovers and, when required, offloads tasks to suitable base stations based on resource availability and network conditions.

\subsubsection{Baseline System}

Baseline represents an established but non-slice-aware IoMT deployment in which heterogeneous applications compete for shared computing and network resources without network slicing or resource-aware orchestration. The baseline system uses a random deployment strategy without any intelligent orchestration framework and network or resource slicing approach. Applications are assigned to base stations without considering their QoS requirements, computational demand, or network conditions. There is no application profiling or resource awareness in the baseline model, which leads to non-optimal placement decisions and higher chances of network congestion or resource underutilization. However, to ensure fair comparison, both the baseline solution and the proposed framework use identical simulation topology, computational capacity of BSs, user mobility model, and workload characteristics.

\section{RESULTS AND EVALUATION}
\label{sec:results}

\subsection{ Resource Slicing Behavior in 5G and Beyond Network}
\begin{figure}[!ht]
\centerline{\includegraphics[width=1\linewidth, height =5cm]{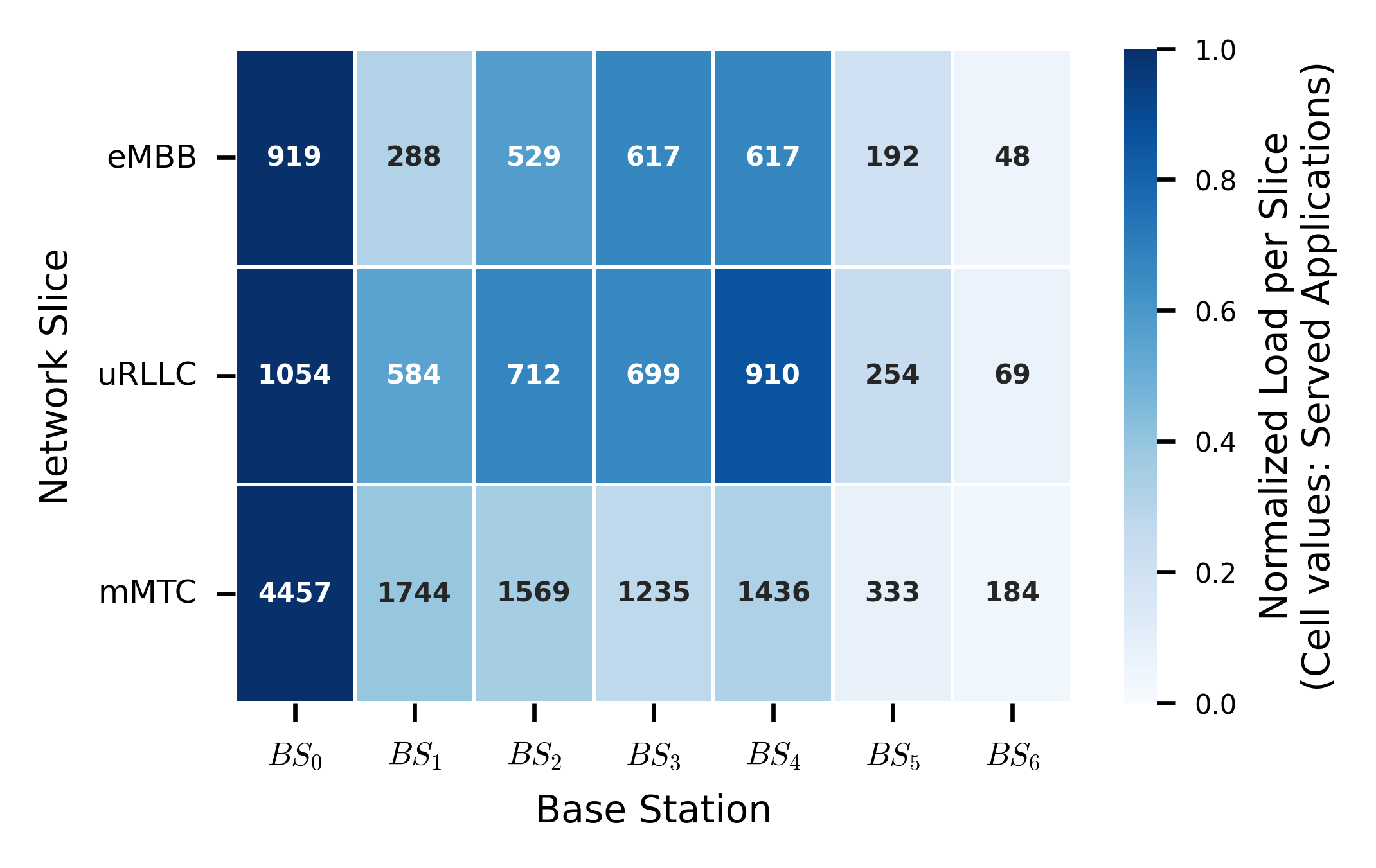}}
\caption{Cumulative served applications per network slice across base stations.}
\label{Served_Applications}
\end{figure}

The behavior of different 5G network slices demonstrates how the proposed system allocates resources based on application requirements. Fig. ~\ref{Served_Applications} shows that the mMTC slice serves the highest number of applications, ranging from 184 to 4457 applications per base station, reflecting its support for massive IoT devices with periodic and lightweight data transmissions. The uRLLC slice serves a moderate number of applications (69–1054), supporting latency-critical services such as XR-assisted surgery, where reliability and delay constraints are stringent. In contrast, the eMBB slice serves fewer applications (48–919) due to the higher demand for data-intensive services, such as doctor-on-demand applications. Overall, the results confirming that the proposed Resource Slicing mechanism effectively isolates traffic, distributes load across base stations, and maintains QoS for heterogeneous IoMT applications.

\subsection{User Mobility Behaviour Across the Regions}
The support for user mobility significantly impacts service continuity in the system. As users move across different base stations coverage areas, frequent handover events are triggered to maintain connectivity. Higher handover activity is observed in central base stations located in region 1 due to higher user density and overlapping coverage areas with other BS, while BS located in the region 2 and 3 experience fewer transitions.

\begin{table}[htbp]
\centering
\caption{Base station handover events during user movement.}
\label{handover_comparison}

\renewcommand{\arraystretch}{1.1}
\setlength{\tabcolsep}{3pt}
\resizebox{\columnwidth}{!}{%
\scriptsize
\begin{tabular}{|c|cc|cc|cc|cc|cc|cc|cc|}
\hline

\multirow{2}{*}{\textbf{From/To}}
& \multicolumn{2}{c|}{\textbf{BS$_0$}} 
& \multicolumn{2}{c|}{\textbf{BS$_1$}}
& \multicolumn{2}{c|}{\textbf{BS$_2$}}
& \multicolumn{2}{c|}{\textbf{BS$_3$}}
& \multicolumn{2}{c|}{\textbf{BS$_4$}}
& \multicolumn{2}{c|}{\textbf{BS$_5$}}
& \multicolumn{2}{c|}{\textbf{BS$_6$}} \\
\cline{2-15}

& \textbf{P} & \textbf{B}
& \textbf{P} & \textbf{B}
& \textbf{P} & \textbf{B}
& \textbf{P} & \textbf{B}
& \textbf{P} & \textbf{B}
& \textbf{P} & \textbf{B}
& \textbf{P} & \textbf{B} \\
\hline

\textbf{BS$_0$}
& 0 & 0
& 31 & 32
& 29 & 28
& 32 & 30
& 38 & 42
& 0 & 0
& 0 & 0 \\
\hline

\textbf{BS$_1$}
& 18 & 23
& 0 & 0
& 0 & 0
& 0 & 0
& 19 & 12
& 0 & 0
& 6 & 6 \\
\hline

\textbf{BS$_2$}
& 19 & 15
& 0 & 0
& 0 & 0
& 22 & 16
& 1 & 0
& 10 & 7
& 0 & 0 \\
\hline

\textbf{BS$_3$}
& 25 & 21
& 0 & 0
& 13 & 13
& 0 & 0
& 0 & 0
& 0 & 0
& 0 & 0 \\
\hline

\textbf{BS$_4$}
& 25 & 27
& 13 & 9
& 1 & 1
& 0 & 0
& 0 & 0
& 0 & 0
& 0 & 0 \\
\hline

\textbf{BS$_5$}
& 0 & 0
& 0 & 0
& 2 & 5
& 0 & 0
& 0 & 0
& 0 & 0
& 0 & 0 \\
\hline

\textbf{BS$_6$}
& 0 & 0
& 4 & 5
& 0 & 0
& 0 & 0
& 0 & 0
& 0 & 0
& 0 & 0 \\
\hline

\end{tabular}
}
\vspace{1mm} \\
\footnotesize{P = Proposed system, B = Baseline system}
 
\end{table}

Compared to the baseline, the proposed approach results in more handover counts, as shown in Table \ref{handover_comparison}, since decisions are guided by resource availability and service requirements. This leads to improved service continuity for real-time IoMT applications requiring uninterrupted operation during mobility.

\subsection{Computational and Resource Usage Analysis}
The proposed framework achieves balanced utilization of computing resources across all base stations. Central nodes handle higher workloads due to increased user density, while BS associated with regions 2 and 3 remain lightly loaded. This distribution reflects realistic edge network behavior.

\begin{figure}[!ht]
\centerline{\includegraphics[width=1\linewidth, height =7cm]{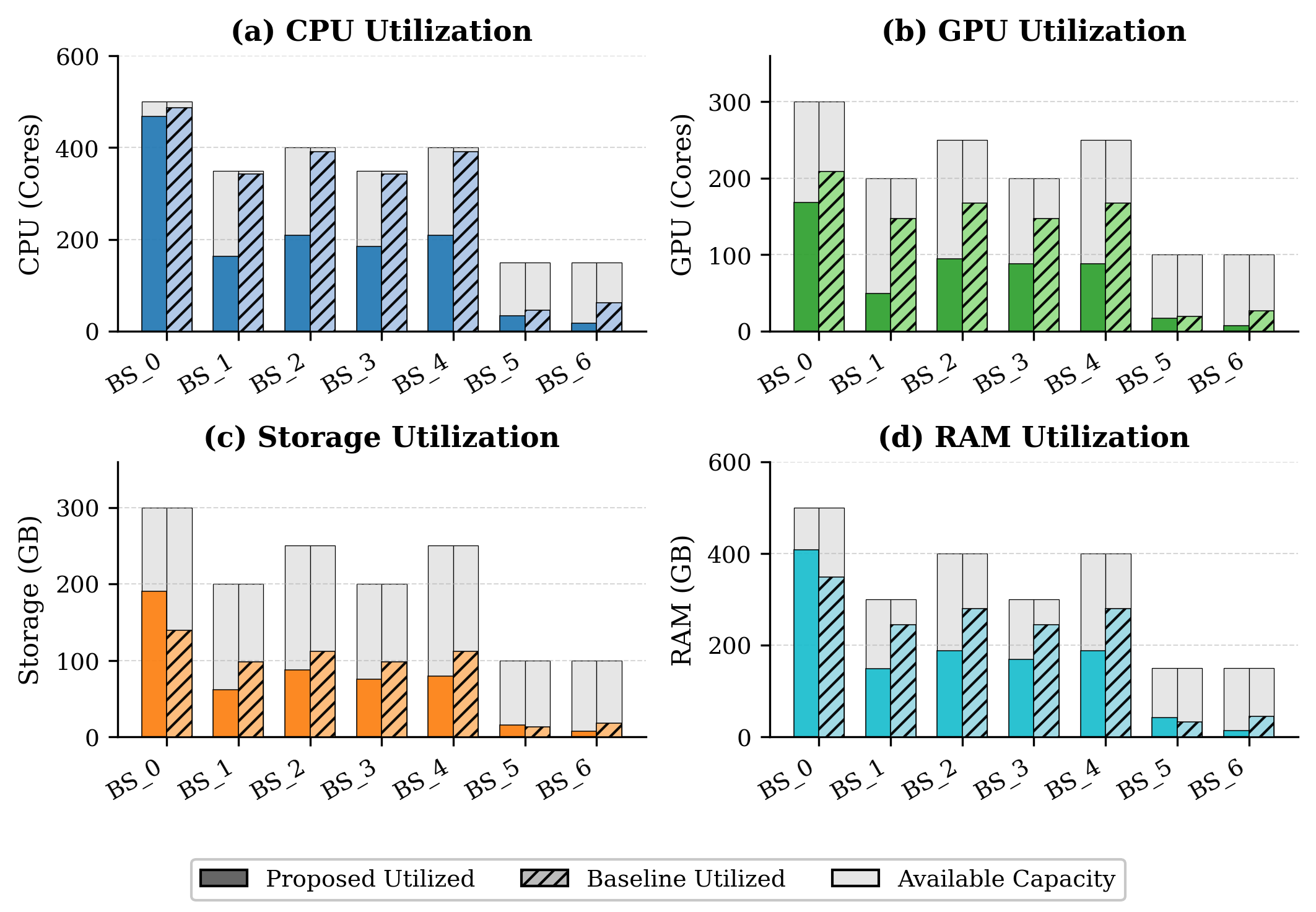}}
\caption{ Comparison of average resource utilization between the proposed and baseline systems for (a) CPU, (b) GPU, (c) Storage, and (d) RAM.}
\label{Resource_utilization_both_systems}
\end{figure}

In contrast to using the baseline system, which often leads to uneven resource usage and early congestion at heavily loaded nodes, the proposed orchestration framework distributes tasks based on both slice requirements and available compute capacity (CPU, RAM, GPU, and storage). As demonstrated in Fig. \ref{Resource_utilization_both_systems}, the proposed system avoids overload situations while significantly improving resource efficiency across the network.

\subsection{System Performance Comparative Analysis}
\begin{figure}[!ht]
\centerline{\includegraphics[width=1\linewidth, height =7cm]{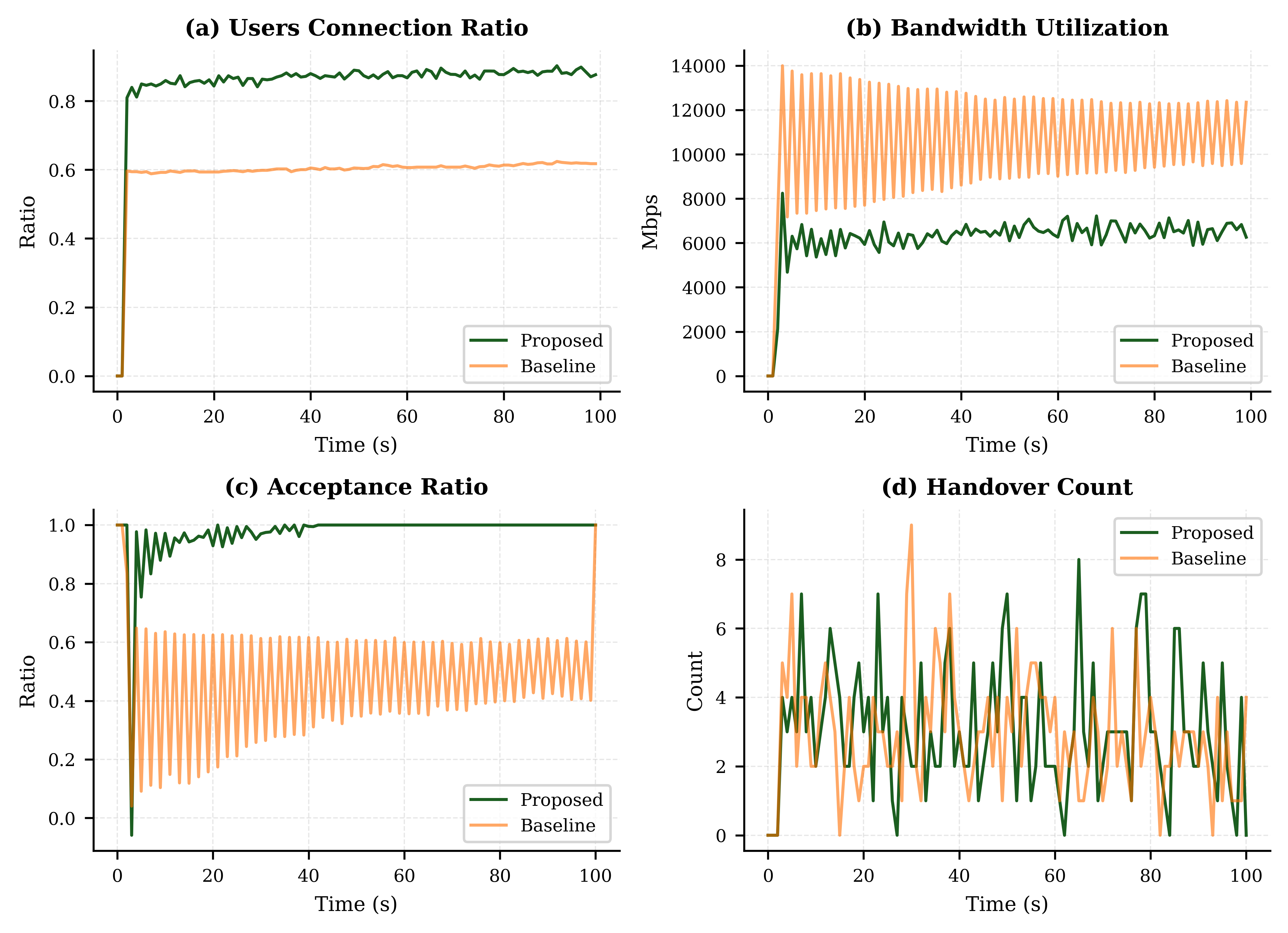}}
\caption{Performance metrics: (a) Connectivity ratio, (b) Bandwidth utilization, (c) Acceptance ratio, and (d) Handover count.}
\label{Performance_Metrics}
\end{figure}

The proposed Resource Slicing framework achieves an acceptance ratio of 0.97 vs 0.47 baseline, and reduces blocking from 0.53 to 0.03.
As shown in Fig. \ref{Performance_Metrics}, bandwidth consumption is also reduced due to efficient task placement and avoidance of unnecessary data transmission. 
Furthermore, the proposed system demonstrates better support for mobility, a higher handover count in the proposed approach is a positive indicator, ensuring seamless service continuity during mobility.

\begin{table}[htbp]
\centering
\caption{COMPARISON OF NETWORK PERFORMANCE METRICS.}
\resizebox{0.48\textwidth}{!}{%
\tiny
\begin{tabular}{|c|c|c|}
\hline
\textbf{Network Metrics} & \textbf{Proposed} & \textbf{Baseline} \\
\hline
Total Users & 500 & 500\\
\hline
Average Connectivity Ratio & 0.85& 0.59 \\
\hline
Average BW Usage (Mbps) & 6241 & 10504 \\
\hline
Average Service Acceptance Ratio & 0.97 & 0.47 \\
\hline
Average Handover Count & 3.08 & 2.88 \\
\hline
\end{tabular}%
}%
\label{Key Network Metrics}
\end{table}

Across all evaluated metrics, the proposed Resource Slicing framework surpasses the baseline as shown in Table \ref{Key Network Metrics}, including service reliability, resource efficiency, and QoS satisfaction. This confirms that combining slice-aware deployment with compute-level resource management improves the performance of distributed IoMT systems in edge environments.

\section{CONCLUSION AND FUTURE WORK}

In this work, we proposed a novel resource slicing-based adaptive orchestration concept, that integrates network slicing with resource-aware service deployment and mobility-aware task offloading for IoMT edge environments. By jointly considering application QoS requirements, network slice constraints and heterogeneous edge resources, the proposed framework improves load distribution, reduces resource congestion, and enhances service reliability compared to a non-slice-aware baseline approach. 
These improvements directly benefit healthcare applications. For XR-assisted surgery, lower latency and higher reliability ensure smooth and reliable real-time 3D visualization. For wearable monitoring, improved resource utilization enables scalable patient tracking under high device density. For doctor-on-demand services, higher service acceptance and lower blocking enhance real-time video consultation. 

To extend our proposed methodology, future work will compare the proposed framework with state-of-the-art AI-driven orchestration algorithms and real-world IoMT-Edge implementations. However, we aim to address real-world deployment challenges, including network failures, security threats, and energy constraints. AI-assisted and predictive orchestration techniques will be investigated to enable proactive resource allocation, intelligent service migration, and mobility prediction. Furthermore, the framework can further be extended to next-generation 6G networks with ultra-low latency and highly reliable communication requirements.

\section*{Acknowledgment}

This work was supported by the 6G Flagship (Grant Number 369116) and 6GESS (Profi6 Grant Number 336449) funded by the Research Council of Finland.

\bibliographystyle{ieeetr}
\bibliography{IEEEexample}

@article{islam2025hybrid,
  title={A hybrid fog-edge computing architecture for real-time health monitoring in IoMT systems with optimized latency and threat resilience},
  author={Islam, Umar and Alatawi, Mohammed Naif and Alqazzaz, Ali and Alamro, Sulaiman and Shah, Babar and Moreira, Fernando},
  journal={scientific reports},
  volume={15},
  number={1},
  pages={25655},
  year={2025},
  publisher={Nature Publishing Group UK London}
}

@article{aguru2022integrated,
  title={Integrated industrial reference architecture for smart healthcare in internet of things: a systematic investigation},
  author={Aguru, Aswani Devi and Babu, Erukala Suresh and Nayak, Soumya Ranjan and Sethy, Abhisek and Verma, Amit},
  journal={Algorithms},
  volume={15},
  number={9},
  pages={309},
  year={2022},
  publisher={MDPI}
}

@ARTICLE{11524197,
  author={Shahid, Hafiz Faheem and Zeeshan, Muhammad and Harjula, Erkki},
  journal={IEEE Access}, 
  title={Autonomous Orchestration of IoMT Applications With Network Slicing in Distributed Edge Computing}, 
  year={2026},
  volume={14},
  number={},
  pages={76601-76615},
  doi={10.1109/ACCESS.2026.3694755}}

@inproceedings{shahid2024resource,
  title={Resource slicing through intelligent orchestration of energy-aware iot services in edge-cloud continuum},
  author={Shahid, Hafiz Faheem and Harjula, Erkki},
  booktitle={Proceedings of the 14th International Conference on the Internet of Things},
  pages={244--245},
  year={2024}
}

@ARTICLE{11392773,
  author={Shahid, Hafiz Faheem and Akdemir, Bilgehan and Islam, Johirul and Ahmad, Ijaz and Ahmad, Ijaz and Harjula, Erkki},
  journal={IEEE Internet of Things Journal}, 
  title={IoT Service Orchestration in Edge–Cloud Continuum With 6G: A Review}, 
  year={2026},
  volume={13},
  number={10},
  pages={20339-20358},
  doi={10.1109/JIOT.2026.3663918}}

@inproceedings{selem2019health,
  title={E-Health applications over 5G networks: challenges and state of the art},
  author={Selem, Enas and Fatehy, Mohammed and Abd El-Kader, Sherine M},
  booktitle={2019 6th International Conference on Advanced Control Circuits and Systems (ACCS) \& 2019 5th International Conference on New Paradigms in Electronics \& information Technology (PEIT)},
  pages={111--118},
  year={2019},
  organization={IEEE}
}

@article{ajakwe2022key,
  title={Key wearable device technologies parameters for innovative healthcare delivery in B5G network: A review},
  author={Ajakwe, Simeon Okechukwu and Nwakanma, Cosmas Ifeanyi and Kim, Dong-Seong and Lee, Jae-Min},
  journal={IEEE Access},
  volume={10},
  pages={49956--49974},
  year={2022},
  publisher={IEEE}
}

@INPROCEEDINGS{10287519,
  author={Lovén, Lauri and Shahid, Hafiz Faheem and Nguyen, Le and Harjula, Erkki and Silvén, Olli and Pirttikangas, Susanna and López, Miguel Bordallo},
  booktitle={2023 20th Annual IEEE International Conference on Sensing, Communication, and Networking (SECON)}, 
  title={Semantic Slicing across the Distributed Intelligent 6G Wireless Networks}, 
  year={2023},
  volume={},
  number={},
  pages={79-84},
  doi={10.1109/SECON58729.2023.10287519}
  }

@article{suleiman2023telemedicine,
  title={Telemedicine and smart healthcare—the role of artificial intelligence, 5G, cloud services, and other enabling technologies},
  author={Suleiman, Taofik Ahmed and Adinoyi, Abdulkareem},
  journal={International Journal of Communications, Network and System Sciences},
  volume={16},
  number={3},
  pages={31--51},
  year={2023},
  publisher={Scientific Research Publishing, Inc.}
}

@ARTICLE{9272281,
  author={Banchs, Albert and de Veciana, Gustavo and Sciancalepore, Vincenzo and Costa-Perez, Xavier},
  journal={IEEE Access}, 
  title={Resource Allocation for Network Slicing in Mobile Networks}, 
  year={2020},
  volume={8},
  number={},
  pages={214696-214706},
  doi={10.1109/ACCESS.2020.3040949}
  }

@article{bahramisirat2024multi,
  title={Multi-access edge computing resource slice allocation: A review},
  author={Bahramisirat, Faezeh and Gregory, Mark A and Li, Shuo},
  journal={IEEE Access},
  volume={12},
  pages={188572--188589},
  year={2024},
  publisher={IEEE}
}

@article{hwang2021iot,
  title={IoT service slicing and task offloading for edge computing},
  author={Hwang, Jaeyoung and Nkenyereye, Lionel and Sung, Nakmyoung and Kim, Jaeho and Song, Jaeseung},
  journal={IEEE Internet of Things journal},
  volume={8},
  number={14},
  pages={11526--11547},
  year={2021},
  publisher={IEEE}
}

@inproceedings{husain2018mobile,
  title={Mobile edge computing with network resource slicing for Internet-of-Things},
  author={Husain, Syed and Kunz, Andreas and Prasad, Athul and Samdanis, Konstantinos and Song, JaeSeung},
  booktitle={2018 IEEE 4th World Forum on Internet of Things (WF-IoT)},
  pages={1--6},
  year={2018},
  organization={IEEE}
}

@INPROCEEDINGS{11220483,
  author={Shahid, Hafiz Faheem and Islam, Johirul and Ahmad, Ijaz and Harjula, Erkki},
  booktitle={2025 IEEE Intelligent Mobile Computing (MobileCloud)}, 
  title={Optimizing Resource-Aware Service Orchestration in Edge-Cloud Continuum}, 
  year={2025},
  volume={},
  number={},
  pages={44-50},
  doi={10.1109/MobileCloud66020.2025.00011}}

@inproceedings{natalino2018machine,
  title={Machine learning aided orchestration in multi-tenant networks},
  author={Natalino, Carlos and Raza, Muhammad Rehan and Rostami, Ahmad and {\"O}hlen, Peter and Wosinska, Lena and Monti, Paolo},
  booktitle={2018 IEEE Photonics Society Summer Topical Meeting Series (SUM)},
  pages={125--126},
  year={2018},
  organization={IEEE}
}

@inproceedings{shahidswarm,
author = {Shahid, Hafiz Faheem and Pahl, Claus},
year = {2019},
month = {09},
pages = {},
title = {Enhanced Particle Swarm Optimisation and Multi Objective Optimization for the Orchestration of Edge Cloud Clusters}
}

@inproceedings{pydi2020analytical,
  title={Analytical review and study on load balancing in edge computing platform},
  author={Pydi, Harikrishna and Iyer, Ganesh Neelakanta},
  booktitle={2020 Fourth international conference on computing methodologies and communication (ICCMC)},
  pages={180--187},
  year={2020},
  organization={IEEE}
}

@INPROCEEDINGS{10503485,
  author={Shahid, Hafiz Faheem and Harjula, Erkki},
  booktitle={2024 IEEE International Conference on Pervasive Computing and Communications Workshops and other Affiliated Events (PerCom Workshops)}, 
  title={Resource-aware Orchestration of IoT Applications in Edge-Cloud Continuum with 6G}, 
  year={2024},
  volume={},
  number={},
  pages={245-246},
  doi={10.1109/PerComWorkshops59983.2024.10503485}}

\end{document}